\documentclass[preprint,review,12pt]{elsarticle}

\usepackage{amssymb}

\journal{Pattern Recognition}

\usepackage[figuresright]{rotating}

\usepackage{times}
\usepackage{soul}
\usepackage{url}
\usepackage[hidelinks]{hyperref}
\usepackage[utf8]{inputenc}
\usepackage[small]{caption}
\usepackage{amsthm}
\usepackage{amsmath}
\usepackage{cases}
\usepackage{diagbox}
\usepackage{booktabs}
\usepackage{graphicx}
\usepackage{subfigure}
\usepackage{stfloats}
\usepackage{algorithm}
\usepackage{algpseudocode}
\usepackage{amsmath}
\usepackage{graphics}
\usepackage{epsfig}
\usepackage{bm}
\usepackage{amssymb,amsfonts}
\usepackage{color}
\usepackage{float}
\usepackage{lineno}
\usepackage{array}

\usepackage{wrapfig} 
\usepackage{flushend} 

\usepackage{url,multirow}

\usepackage{color}

\begin{document}

\begin{frontmatter}



\title{Embedded Multi-label Feature Selection via Orthogonal Regression}


\address[label1]{Faculty of Information Technology, Beijing University of Technology, Beijing, 100124, CN}
\address[label2]{School of Artificial Intelligence, Beijing Normal University, Beijing, 100875, CN}
\address[label4]{Guangdong Artificial Intelligence and Digital Economy Laboratory, Guangzhou, 511442, CN}
\address[label5]{School of Computer Science and School of Artificial Intelligence, OPTIMAL, Northwestern Polytechnical University, Xian, 710072, CN}

\author[label1,label2,label4]{Xueyuan Xu}
\author[label2]{Fulin~Wei}
\author[label2]{Tianyuan Jia}
\author[label1]{Li~Zhuo}
\author[label5]{Feiping~Nie}
\author[label2,label4]{Xia~Wu\corref{cor1}}

\cortext[cor1]{Corresponding author}
\ead{wuxia@bnu.edu.cn}


\begin{abstract}
In the last decade, embedded multi-label feature selection methods, incorporating the search for feature subsets into model optimization, have attracted considerable attention in accurately evaluating the importance of features in multi-label classification tasks. Nevertheless, the state-of-the-art embedded multi-label feature selection algorithms based on least square regression usually cannot preserve sufficient discriminative information in multi-label data. To tackle the aforementioned challenge, a novel embedded multi-label feature selection method, termed global redundancy and relevance optimization in orthogonal regression (GRROOR), is proposed to facilitate the multi-label feature selection. The method employs orthogonal regression with feature weighting to retain sufficient statistical and structural information related to local label correlations of the multi-label data in the feature learning process. Additionally, both global feature redundancy and global label relevancy information have been considered in the orthogonal regression model, which could contribute to the search for discriminative and non-redundant feature subsets in the multi-label data. The cost function of GRROOR is an unbalanced orthogonal Procrustes problem on the Stiefel manifold. A simple yet effective scheme is utilized to obtain an optimal solution. Extensive experimental results on ten multi-label data sets demonstrate the effectiveness of GRROOR.

\end{abstract}




\end{frontmatter}


\section{Introduction}
\label{}

In many scenarios, an instance can be naturally annotated with multiple semantic labels. For example, an image can be attached with multiple scenes. These scenarios can be treated as multi-label learning tasks. Multi-label learning tasks have attracted significant interest in a variety of practical applications in the field of machine learning and pattern recognition, such as image classification \cite{LI2022108521}, vertebrae identification \cite{qin2020VertebraeLabeling}, and affective state recognition \cite{min2023finding}.

With the increasing growth of feature dimensionality, the performance of the multi-label learning tasks is confronted with the negative impacts of irrelevant and noisy features \cite{zhang2020multi,zhang2023mfsjmi}. To deal with the above curse of feature dimensionality, feature selection has been employed in discarding irrelevant and noisy features while retaining discriminative features \cite{lee2017scls}. The advantage of the feature selection in multi-label learning tasks is that it can preserve the intuitive meaning and physical interpretation, reduce the cost of storage, avoid the curse of dimensionality, and prevent overfitting \cite{gao2023unified}.

Recently, plenty of multi-label feature selection methods have been designed for eliminating irrelevant and noisy features in the feature representation. In accordance with the different searching strategies, current multi-label feature selection methods can be roughly divided into three models: filter, wrapper, and embedded models \cite{li2023robust}. Filter models search feature subsets on the basis of certain characteristics in the multi-label data, such as the Pearson correlation coefficient between each feature and the corresponding label. Wrapper models seek out candidate feature subsets by random or sequential search and then evaluate the fitness of the candidate feature subsets by the performance of the subsequent learning algorithm \cite{tang2014feature}. Although the wrapper models usually have an effective performance, they pay expensive time costs in practical applications, especially when the number of features in the candidate feature subset is large \cite{zhang2019review}.

Different from the search strategy of filter and wrapper models, embedded models directly incorporate the search for feature subsets into the optimization problem \cite{zhang2019manifold}. The final feature subset is obtained by optimizing the objective function of the learning model, which can accurately evaluate the importance of each feature in the performance of the learning model \cite{li2023multi}. Additionally, embedded methods usually have rather lower computational costs than wrapper methods \cite{cai2018feature,fan2024learning}. Due to its completeness in statistical theory and simplicity for data analysis, least square regression is applied as a fundamental statistical analysis technique in the learning model construction of most embedded multi-label feature selection methods \cite{zhang2019review}. Least square regression(LSR)-based multi-label feature selection methods learn a projection matrix $W$ with sparsity restriction by minimizing regression error and the score of each feature is calculated by $\{\|{w}_{1}\|_{2},..., \|{w}_{d}\|_{2}\}$ \cite{cai2013sfs}.

\begin{table}[!t]
\footnotesize
\setlength{\abovecaptionskip}{0.cm}
\setlength{\belowcaptionskip}{-0.cm}
\caption{The difference between previous methods and GRROOR.}\label{tab:difference}
\begin{center}
{
\begin{tabular}{lccccc}
\hline\hline
\multicolumn{1}{l}{\multirow{2}*{Methods}}        &Global       & Local     & Global   & Feature   & Orthogonal \\
&label relevance       &label correlations     & feature redundancy   & weighting   & regression
\\\hline
MIFS         & $\times$          &$\checkmark$        & $\times$          & $\times$      & $\times$  \\
SCLS         & $\times$          &$\checkmark$        & $\times$          & $\times$      & $\times$  \\
SCMFS         & $\checkmark$           & $\times$       & $\times$        & $\times$         & $\times$\\
MDFS           & $\checkmark$          & $\checkmark$      & $\times$     & $\times$           & $\times$   \\
GRRO           & $\checkmark$          & $\times$       & $\checkmark$    & $\times$          & $\times$ \\
MFS\_MCDM         & $\times$          & $\checkmark$      & $\times$      & $\times$          & $\times$   \\
\textbf{GRROOR}           & $\checkmark$          & $\checkmark$       & $\checkmark$    &$\checkmark$           & $\checkmark$ \\
\hline\hline
\end{tabular}}
\end{center}
\end{table}

However, existing LSR-based multi-label feature selection methods have the following limitation. LSR could not preserve sufficient discriminative properties in the projection subspace \cite{wu2020supervised}, which may result in non-optimal feature subsets for the multi-label feature selection task. To tackle the aforementioned challenge, in this paper, we propose a novel embedded multi-label feature selection method via global redundancy and relevance optimization in orthogonal regression (GRROOR). The LSR model can be restricted to the Stiefel manifold, which introduces the orthogonal constraints into the LSR model. Instead of minimizing the horizontal distance in the LSR, orthogonal regression aims to minimize the perpendicular distance from the data points to the regression line. Through the above distance calculation approach, the orthogonal regression could explore more local structural information in the projection subspace \cite{leng2007ordinary}. Then, global feature redundancy information and global label relevance information are both added into the orthogonal regression model to accurately exploit feature redundancy and label relevance from a global view. Table~\ref{tab:difference} comprehensively compare the proposed method with several state-of-the-art multi-label feature selection methods from four different aspects.

Moreover, the major contributions of our work are summarized as follows:

\begin{enumerate}
  \item[$\bullet$] Compared with the state-of-the-art LSR-based multi-label feature selection methods, we employ the orthogonal regression with feature weighting as a novel statistical model for multi-label feature selection. The orthogonal regression can retain more statistical and structural information related to local label correlations in the projection subspace. Each scale factor in the feature weighting matrix is utilized to accurately analyze the importance of the corresponding feature on the multi-label learning task.
  \item[$\bullet$] Global feature redundancy information is introduced into the orthogonal regression-based multi-label feature selection framework to discard redundant features. Then, global label relevance information is also incorporated into the multi-label projection space to explore the label relevance in the multiple labels from a global view and obtain informative and representational low-dimensional label subspace.
  \item[$\bullet$] The objective function of GRROOR is an unbalanced orthogonal Procrustes problem on the Stiefel manifold. To solve the optimization problem of GRROOR, an efficient alternative scheme is developed to ensure convergence and obtain an optimal solution. Extensive experimental evaluation is conducted on ten benchmark multi-label data sets to demonstrate the superiority of the proposed GRROOR method in contrast with nine state-of-the-art multi-label feature selection methods.
\end{enumerate}

The remainder of this paper is organized as follows. Section~\ref{RW} explains the notations and reviews the related researches. Section~\ref{Method} describes the proposed multi-label feature selection framework in detail. We propose an optimization scheme to solve the proposed method in Section~\ref{Optimization Strategy}. The details of data sets, evaluation metrics, experimental setting, and experimental results are introduced in Section~\ref{Experimental Details}. Finally, we conclude this paper in Section~\ref{Conclusions}.

\begin{table}[!b]
\footnotesize
\setlength{\tabcolsep}{18pt}
\setlength{\abovecaptionskip}{0.cm}
\setlength{\belowcaptionskip}{-0.cm}
\caption{Notations}\label{tab:notations}
\begin{center}
{
\begin{tabular}{cc}
\hline\hline
Notation     & Definition  \\
\hline\hline
$d$     & The number of features   \\        
$n$     & The number of samples    \\      
$k$     & The number of classes    \\       
$c$     & The number of clusters   \\   
$\lambda$, $\alpha$, $\beta$, $\eta$    & The balance parameters \\          
$\bm{b}\in {\mathbb{R}^{c \times 1}}$    & A bias vector      \\     
$\bm{\theta}\in {\mathbb{R}^{d \times 1}}$                    &A feature score vector  \\ 
$\bm{x}_{i}\in {\mathbb{R}^{\text{1}\times n}}$	      & The $i$-th feature \\        
$ \bm{1}_n =(1,1, \ldots, 1)^{T} $     & A row vector of all ones      \\       
$X=[\bm{x}_1, \bm{x}_2, ..., \bm{x}_d]^{T}\in {\mathbb{R}^{d\times n}}$   & The feature data matrix  \\
$Y\in {\mathbb{R}^{n\times k}}$                           & The multi-label matrix         \\
 $V\in {\mathbb{R}^{n\times c}}$                           & The latent semantics matrix         \\
$B\in {\mathbb{R}^{c\times k}}$                           & The coefficient matrix         \\
 $I_n\in {\mathbb{R}^{n\times n}}$     & An $n \times n$ identity matrix            \\
 $W\in {\mathbb{R}^{d \times c}}$                                    & An orthogonal matrix            \\
 $\Theta \in {\mathbb{R}^{d\times d}}$     & A diagonal matrix            \\
$A\in {\mathbb{R}^{d\times d}}$    & A feature redundancy matrix            \\
$\|.\|_{F}$    & The Frobenius norm of a matrix            \\
$vec$    & The vectorization of a matrix           \\
$t r\left(.\right)$    & The trace of a square matrix            \\
\hline\hline
\end{tabular}}
\end{center}
\end{table}

\section{Related Works}\label{RW}
\subsection{Notations and Definitions}
Throughout the full text, vectors and matrices are denoted by lowercase boldface letters ($\bm{a}$, $\bm{b}$, ...) and uppercase letters ($A$, $B$, ...), respectively. The operators $\circ$ and $vec$ are the Hadamard product and vectorization. The transposition and trace of a matrix are represented by uppercase superscript $T$ and $tr$. $ \bm{1}_n =(1, \ldots, 1)^{T} \in \mathbb{R}^{n \times 1}$. $I_n$ represents an $n \times n$ identity matrix. Notations are summarized in Table~\ref{tab:notations}.

Given a multi-label data set ($X$,$Y$), $X=[\bm{x}_1, \bm{x}_2, ..., \bm{x}_d]^{T}\in {\mathbb{R}^{\text{d}\times n}}$ is the data matrix where $\bm{x}_{d}\in {\mathbb{R}^{\text{1}\times n}}$, and $Y=\left[\bm{y}_{.1}, \bm{y}_{.2}, \ldots, \bm{y}_{.k}\right] \in\{-1,1\}^{n\times k}$ is the multi-label matrix where $i$-th label $\bm{y}_{.i}=\left\{y_{1i}, \ldots, y_{ni}\right\}^{T} \in\{-1,1\}^{n\times 1}$. $d$, $n$, and $k$ are the number of features, samples, and labels, respectively.

The Frobenius norm of a matrix $S$ is denoted as:
\begin{equation}
\|S\|_{F}=\sqrt{\sum_{i=1}^{m} \sum_{j=1}^{n} s_{i j}^{2}}=\sqrt{t r\left(S^{T} S\right)}
\end{equation}

The $l_{2,1}$-norm of $S$ is denoted as:
\begin{equation}
\|S\|_{2,1}=\sum_{i=1}^{m} \sqrt{\sum_{j=1}^{n} s_{i, j}^{2}}=\sum_{i=1}^{n}\left\|s_{i}\right\|_{2}
\end{equation}

\subsection{A review of embedded multi-label feature selection methods}
Embedded methods embed the feature selection process into the model optimization and rank the feature importance in the performance of multi-label learning, such as multi-label informed feature selection (MIFS) \cite{jian2016multi}, learning label-specific features (LLSF)\cite{huang2016learning}, manifold-based constraint Laplacian score (MCLS) \cite{huang2018manifold}, multi-label learning with global and local label correlation \cite{zhu2018GLOCAL}, embedded feature selection method via manifold regularization (MDFS) \cite{zhang2019manifold}, shared common mode between features and labels (SCMFS)\cite{hu2020multi}, and multi-label feature selection using multi-criteria decision making (MFS-MCDM) \cite{hashemi2020mfs}.

To perform feature selection, the majority of the above-embedded models implement sparse constraints to the projection matrix, including $l_{1}$-norm, $l_{2}$-norm, and $l_{2,1}$-norm. For example, LLSF is a $l_{1}$-norm regularized least square regression mode for embedded multi-label feature selection.  The objective function of LLSF is defined as follows:

\begin{equation}
\min _{W} \frac{1}{2}\|X W-Y\|_F^2+\frac{\alpha}{2} \operatorname{Tr}\left(R {W}^T W\right)+\beta\|W\|_1,
\end{equation}

To choose discriminative features that are shared by multiple labels, motivated by LSI \cite{dumais2004latent}, MIFS \cite{jian2016multi} exploited label correlations by projecting the high-dimensional multi-label space $Y$ into a low-dimensional label subspace $V$. The framework of MIFS is represented as
\begin{equation}
\label{equ_mifs}
\begin{aligned}
\min _{W, V, B} \|X^{T} W-V\|_{F}^{2}+\alpha\|Y-V B\|_{F}^{2}+\beta \operatorname{Tr}\left(V^{T} L V\right)+\gamma\|W\|_{2,1} \\
\end{aligned}
\end{equation}
where $V\in {\mathbb{R}^{\text{n}\times c}}$, $B\in {\mathbb{R}^{\text{c}\times k}}$, and $L\in {\mathbb{R}^{\text{n}\times n}}$ represent the latent semantics of the multiple label information, the coefficient matrix, and the graph Laplacian matrix, respectively. $L=G-S$. $G$ is a diagonal matrix with $G_{i i}=\sum_{j=1}^{n} S_{i j}$. The element $S_{i j}$ in the affinity graph $S$ is the similarity measure of samples $\bm{x}_{.i}$ and $\bm{x}_{.j}$. The affinity graph $S$ can be calculated by a heat kernel.
\begin{equation}
S_{i j}=\left\{\begin{array}{lc}
\exp \left(-\frac{\left\|\bm{x}_{i}-\bm{x}_{j}\right\|^{2}}{\sigma^{2}}\right) &\bm{x}_{i} \in \mathcal{N}_{p}\left(\bm{x}_{j}\right) \text { or } \bm{x}_{j} \in \mathcal{N}_{p}\left(\bm{x}_{i}\right) \\
0 & \text { otherwise }
\end{array}\right.
\end{equation}

The term $\|Y-V B\|_{F}^{2}$ clusters the original $k$ labels into $c$ clusters to capture the semantics in the multiple labels. The term $\operatorname{Tr}\left(V^{T} L V\right)$ tries to guarantee that local geometry structures are consistent between the input feature data $X$ and the low-dimensional label subspace $V$ \cite{jian2016multi}. The strategy of reducing multi-label dimension was adopted in many researches, including, but not limited to, SCMFS \cite{hu2020multi}, DRMFS \cite{hu2020DRMFS}, correlated and multi-label feature selection method (CMFS)\cite{braytee2017multi}.

For example, SCMFS employs CMF to discover the shared common mode information between the feature matrix and the multi-label matrix, taking into account the comprehensive data information in the two matrices. In addition, SCMFS uses non-negative matrix factorization to enhance the interpretability for feature selection \cite{hu2020multi}. The objective function of SCMFS is as follows:

\begin{equation}
\begin{aligned}
\min _{W, V, Q, B}&\|X^{T} W-V\|_{F}^{2}+\alpha\|X^{T}-V Q\|_{F}^{2}+\beta\|Y-V B\|_{F}^{2} +\gamma\|W\|_{2,1} \\
&\text { s.t. }\{W, V, Q, B\} \geq 0
\end{aligned}
\end{equation}
where $Q\in {\mathbb{R}^{\text{c}\times d}}$ is the coefficient matrix of the data matrix $X$. Different from MIFS, the $V$ in SCMFS is the shared common mode between the data matrix $X$ and the label matrix $Y$.

\section{The Proposed Framework}\label{Method}
In this section, the GRROOR framework is illustrated in detail.

\subsection{Problem Formulation}
\begin{figure*}[!t]
\centering
\includegraphics[width=0.995\textwidth]{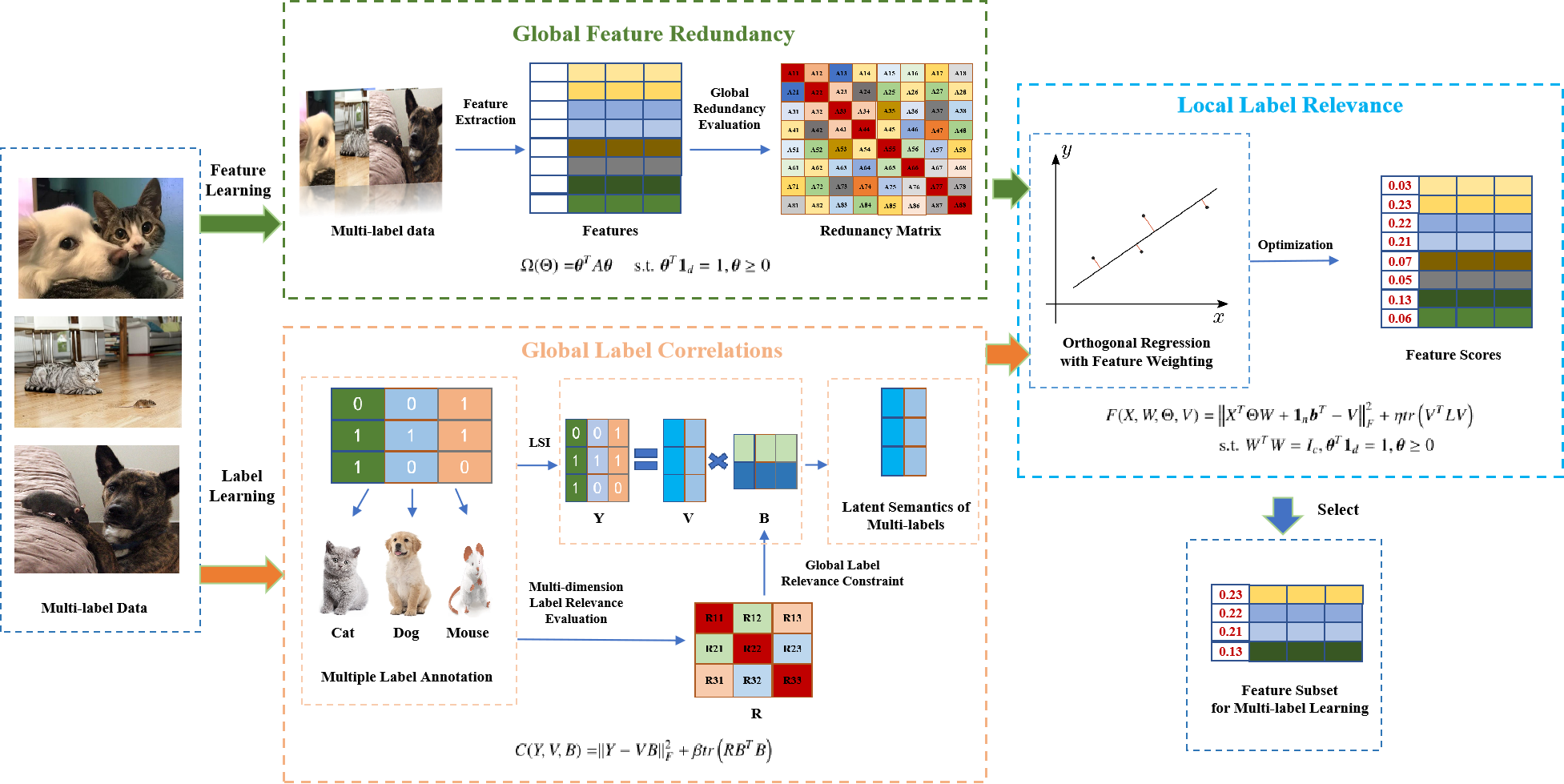}
\caption{The GRROOR framework consists of three sections: (a) exploring  global feature redundancy; (b) exploiting global label correlations; (3) evaluating local label relevance.}
\label{Framework_grroor}
\end{figure*}

To obtain informative and non-redundant feature subsets for the multi-label learning, a novel embedded multi-label feature selection method is proposed in this section. The proposed GRROOR framework is defined as follows:

\begin{equation}
\min _{W, \Theta, V, B} F(X, W, \Theta, V)+\gamma C(Y, V, B)+\lambda \Omega(\Theta)
\end{equation}
where $W$, $\Theta$, $V$, and $B$ are projection matrix, feature weighting matrix, latent semantics of the multiple label information, and coefficient matrix, respectively. $\lambda$ and $\gamma$ represent tradeoff parameters. The terms $F$, $C$, and $\Omega$ denote the feature mapping function, the multi-label learning function, and the feature redundancy function, respectively. Firstly, the feature mapping function is employed to capture the local label correlations between features and labels. Additionally, the multi-label learning function is adopted to exploit the global label relevance. Finally, the feature redundancy function is introduced to mine the redundancy between features from a global view. The proposed GRROOR framework is shown in Fig.~\ref{Framework_grroor}. The detailed definitions of the above terms $F$, $C$, and $\Omega$ will be introduced in the following sections.

\subsection{Explore local label correlations}
To obtain more local structural information in the projection subspace and rank the weights of all original features, orthogonal constraint ${{W}^{T}}W={{I}_{c}}$ and feature weighting $\Theta$ are introduced in the feature mapping function. The term $F$ can be formulated as follows:
\begin{equation}
\begin{aligned}
\label{equ_framework_llc}
F(X, W, \Theta, V) = & \left\|X^{T} \Theta W + \bm{1}_{n} \bm{b}^{T}-V\right\|_{F}^{2} +\eta tr\left(V^{T} L V\right) \\
&\text { s.t. } W^{T} W=I_{c}, \bm{\theta}^{T} \bm{1}_{d}=1, \bm{\theta} \geq 0
\end{aligned}
\end{equation}
where $W\in {\mathbb{R}^{d\times c}}$ with orthogonal constrain ${{W}^{T}}W={{I}_{c}}$ is the orthogonal projection subspace, and $\bm{b}\in {\mathbb{R}^{c\times 1}}$ represents the bias. $\eta$ (${\eta} > 0$) is a tradeoff parameter. Different from least square regression based multi-label feature selection models, a feature score vector $\bm{\theta}\in {\mathbb{R}^{d\times 1}}$ ($\bm{\theta} \ge 0$, ${\bm{\theta}^{T}}{{\bm{1}}_{d}}=1$) is adopted to evaluate the importance of each feature in the multi-label learning tasks. $\Theta \in {\mathbb{R}^{d\times d}}$ is a diagonal matrix with $\Theta_{i i} = \bm{\theta}_{i}$.

Next, the term $\operatorname{Tr}\left(V^{T} L V\right)$ is introduced to retain that the local geometry structures are consistent between the original feature space $X$ and the latent semantics space $V$ \cite{jian2016multi}. $L=G-S$ represents the graph Laplacian matrix ($L\in {\mathbb{R}^{\text{n}\times n}}$). $S$ is the affinity graph of $X$, and $G$ denotes a diagonal matrix with $G_{i i}=\sum_{j=1}^{n} S_{i j}$. The affinity graph $S$ is calculated by a heat kernel. The element $S_{i j}$ in $S$ is the similarity value of two instances $\bm{x}_{.i}$ and $\bm{x}_{.j}$. The definition of $S_{i j}$ is:
\begin{equation}
S_{i j}=\left\{\begin{array}{lc}
\exp \left(-\frac{\left\|\bm{x}_{.i}-\bm{x}_{.j}\right\|^{2}}{\sigma^{2}}\right) &\bm{x}_{.i} \in \mathcal{N}_{p}\left(\bm{x}_{.j}\right) \text { or } \bm{x}_{.j} \in \mathcal{N}_{p}\left(\bm{x}_{.i}\right) \\
0 & \text { otherwise }
\end{array}\right.
\end{equation}
where $\sigma$ and $\mathcal{N}_{p}\left(\bm{x}_{j}\right)$ denote the graph construction parameter and the set of top-$p$ nearest neighbors of the instance $\bm{x}_{.j}$.

\subsection{Exploit global label relevance}
On the basis of the latent semantic indexing mechanism in Eq.~\eqref{equ_mifs}, $\bm{b}_{. i}$ and $\bm{b}_{. j}$ in the coefficient matrix $B$ denote the coefficient of label $\bm{y}_{.i}$ and $\bm{y}_{.j}$ in LSI. If the labels $\bm{y}_{.i}$ and $\bm{y}_{.j}$ are strongly correlated, the clustering results (i.e, $\bm{b}_{. i}$ and $\bm{b}_{. j}$) in the coefficient matrix $B$ should be similar. Otherwise, $\bm{b}_{. i}$ and $\bm{b}_{. j}$ should have a great difference. Hence, the new classification information for the two labels $\bm{y}_{.i}$ and $\bm{y}_{.j}$ can be saved in the label dimension reduction process. To realize the goal, a regularizer for the coefficient matrix $B$ is defined as:
\begin{equation}
\label{equ_label_relevance}
\sum_{i=1}^{k} \sum_{j=1}^{k} R_{i j} \bm{b}_{. i}^{T} \bm{b}_{. j}
\end{equation}
where $R_{i j}=1-Z_{i j}$, and $Z_{i j}$ indicates the relevance between labels $\bm{y}_{.i}$ and $\bm{y}_{.j}$. The global label relevance matrix $Z$ is calculated to guide the latent semantic indexing process $\|Y- V B\|_{F}^{2}$. The elements in $Z$ is computed by cosine similarity to mine second-order correlations among multiple labels. Based on the above analysis, Eq.~\eqref{equ_label_relevance} can be integrating with the latent semantic indexing process to exploit global label relevance in the latent semantics. Hence, the multi-label learning function can be written as the following:

\begin{equation}
\begin{aligned}
\label{equ_framework_R}
C(Y, V, B)= & \|Y- V B\|_{F}^{2} +\beta tr\left(RB^{T}B\right) \\
\end{aligned}
\end{equation}
where $R\in {\mathbb{R}^{k\times k}}$ is employed to exploit global label relevance. $\beta$ (${\beta} > 0$) is a tradeoff parameter. The latent semantic indexing process clusters the original $k$ labels into $c$ clusters to capture the semantics in the $k$ labels. It can be easily seen that $R$ is positive semi-definite.

\subsection{Evaluate global feature redundancy}
Furthermore, a global feature redundancy matrix $A$ is introduced to evaluate the redundancy among all the original features. The elements in $A$ are defined as follows:
\begin{equation}
\label{equ_red_matrix}
A_{i, j}=\left(O_{i, j}\right)^{2}=\left(\frac{\bm{f}_{i}^{T} \bm{f}_{j}}{\left\|\bm{f}_{i}\right\|\left\|\bm{f}_{j}\right\|}\right)^{2}
\end{equation}
where $\bm{f}_{i}\in {\mathbb{R}^{n\times 1}}$ and $\bm{f}_{j}\in {\mathbb{R}^{n\times 1}}$ are $i$-th and $j$-th centralized features of $\bm{x}_{i}$ and $\bm{x}_{j}$ ($i,j=1,2,...,\textit{d}$), respectively. $\bm{f}_{i}$ and $\bm{f}_{j}$ can be computed as
\begin{equation}\left\{\begin{array}{l}
\bm{f}_{i}=H \bm{x}_{i}^{T} \\
\bm{f}_{j}=H \bm{x}_{j}^{T}
\end{array}\right.\end{equation}
where $H={{I}_{\text{n}}}-\frac{1}{n}{{\bm{1}}_{n}}{{\bm{1}}_{n}}^{T}$. Eq.~\eqref{equ_red_matrix} can be reformulated as
\begin{equation}
O=D F^{T} F D=(F D)^{T} F D
\end{equation}
where $F=[\bm{f}_1, \bm{f}_2, ..., \bm{f}_d]$. $D$ denote a diagonal matrix with $D_{i, i}=\frac{1}{\left\|\bm{f}_{i}\right\|}$ ($i=1,2,...,\textit{d}$). The matrix $O$ is positive semi-definite. On the basis of $A = O \circ O$, $A$ is a non-negative and positive semi-definite matrix\cite{wang2015grm}.

To realize the global feature redundancy minimization in the orthogonal regression, a regularizer for the feature score vector $\bm{\theta}$ is added as the following:
\begin{equation}
\begin{aligned}
\label{equ_framework_theta}
\Omega(\Theta)= &  \bm{\theta}^{T} A \bm{\theta}  &\text { s.t. } \bm{\theta}^{T} \bm{1}_{d}=1, \bm{\theta} \geq 0
\end{aligned}
\end{equation}

The term $\bm{\theta}^{T} A \bm{\theta}$ can be written as $\sum_{i, j=1}^{d} A_{i, j} \bm{\theta}_{i} \bm{\theta}_{j}$. $\bm{\theta}_{i}$ and $\bm{\theta}_{j}$ represent the scores of the features $\bm{x}_{i}$ and $\bm{x}_{j}$ evaluated by the term $F$ in Eq.~\eqref{equ_framework_llc}. The large value of $A_{i, j}$ denotes that $\bm{x}_{i}$ and $\bm{x}_{j}$ are dependent. When $\bm{\theta}_{i}> \bm{\theta}_{j}$, the score $\bm{\theta}_{j}$ will be automatically reduced with larger $\bm{\theta}_{i}$ to minimize the value of $\bm{\theta}^{T} A \bm{\theta}$. Videlicet, when $\bm{x}_{i}$ and $\bm{x}_{j}$ are dependent and redundant, the value of corresponding $\bm{\theta}_{i}$ will remain unchanged and that of corresponding $\bm{\theta}_{j}$ will be automatically reduced.

\subsection{The final objective function of GRROOR}
Based on the above analysis, the final objective function of GRROOR is obtained as follows:
\begin{equation}
\begin{aligned}
\label{equ_framework}
\min\limits_{W, \bm{b}, \Theta, V, B}& \left\|X^{T} \Theta W +  \bm{1}_{n} \bm{b}^{T}-V\right\|_{F}^{2}+\alpha\|Y- V B\|_{F}^{2} \\
&+\eta tr\left(V^{T} L V\right)+\lambda \bm{\theta}^{T} A \bm{\theta} +\beta tr\left(RB^{T}B\right) \\
&\text { s.t. } W^{T} W=I_{c}, \bm{\theta}^{T} \bm{1}_{d}=1, \bm{\theta} \geq 0
\end{aligned}
\end{equation}
where $\alpha$, $\eta$, $\lambda$, and $\beta$ denote tradeoff parameters.

More specifically, in Eq.~\eqref{equ_framework}, orthogonal regression is adopted as the statistical analysis model. Compared with least square regression, orthogonal regression could retain more local structural information of multi-label data. Then, the feature weighting matrix $\Theta$ with global redundancy minimization constraint $\bm{\theta}^{T} A \bm{\theta}$ is added into the orthogonal regression model to accurately explore the feature relevance and redundancy from a global view. Lastly, the high-dimensional label space $Y$ is projected into a low-dimensional subspace $V$ with global label relevance optimization constraint $tr\left(RB^{T}B\right)$ to effectively explore global label relevance. By optimizing Eq.~\eqref{equ_framework}, the global redundancy and relevance optimization are realized simultaneously.

\section{Optimization Strategy} \label{Optimization Strategy}
By virtue of the extreme value condition w.r.t $\bm{b}$, we can derive the optimal solution of $\bm{b}$ $\bm{b}=\frac{1}{n}\left(V^{T} \bm{1}_{n}-W^{T} \Theta X \bm{1}_{n}\right)$. Substituting the optimal solution of $\bm{b}$ into Eq.~\eqref{equ_framework}, we can rewrite Eq.~\eqref{equ_framework} as

\begin{equation}
\begin{aligned}
\label{equ_framework_re}
\min\limits_{W, \Theta, V, B}&\left\|H X^{T} \Theta W -H V \right\|_{F}^{2}+\alpha\|Y- V B\|_{F}^{2} \\
&+\eta tr\left(V^{T} L V\right)+\lambda \bm{\theta}^{T} A \bm{\theta} +\beta tr\left(RB^{T}B\right) \\
&\text { s.t. } W^{T} W=I_{c}, \bm{\theta}^{T} \bm{1}_{d}=1, \bm{\theta} \geq 0
\end{aligned}
\end{equation}

We can apply an alternative optimization approach to solve for $W$, $\Theta$, $V$, and $B$ in Eq.~\eqref{equ_framework_re}. The optimization of Eq.~\eqref{equ_framework_re} is further decomposed into the following four subproblems.

\subsection{Update $W$ by fixing $\Theta$, $V$, and $B$}
With the fixed $\Theta$, $V$, and $B$, Eq.~\eqref{equ_framework_re} is formulated as:
\begin{equation}
\label{equ_qpsm}
\min\limits_{W^{T} W=I_{c}} tr\left(W^{T} J W-2 W^{T} M\right)
\end{equation}
where $ J=\Theta XH{{X}^{T}}{{\Theta }^{T}}$ and $M=\Theta XH{{V}}$. Eq.~\eqref{equ_qpsm} is related to the quadratic problem on the Stiefel manifold (QPSM). Generalized power iteration (GPI) method \cite{nie2017GPI} is introduced to address the mathematic issue. Compared with other methods, the GPI algorithm takes lower computation costs and becomes more efficient in dealing with high-dimension data matrices. The specific solution process to $W$ in the GPI algorithm is shown in \cite{nie2017GPI}.

\subsection{Update $\Theta$ by fixing $W$, $V$, and $B$}
With the fixed $W$, $V$, and $B$, the irrelevant items of $\Theta$ are ignored and Eq.~\eqref{equ_framework_re} is rewritten as:
\begin{equation}
\begin{aligned}
\label{equ_framework_re2}
\min\limits_{\Theta} & \left[{tr\left(\Theta X H X^{T} \Theta W W^{T}\right)+ \lambda \bm{\theta}^{T} A \bm{\theta}} \right. \left.{-tr\left(2 \Theta X H V W^{T}\right)}\right]\\
& {\text { s.t. } W^{T} W=I_{c}, \bm{\theta}^{T} \bm{1}_{d}=1, \bm{\theta} \geq 0}
\end{aligned}
\end{equation}

Eq.~\eqref{equ_framework_re2} can be reformulated as follows:
\begin{equation}
\label{equ_solution_theta}
\begin{array}{l}{\min\limits_{\bm{\theta}}\left[\bm{\theta}^{T}\left[\left(X H X^{T}\right)^{T} \circ\left(W W^{T}\right)+\lambda A \right] \bm{\theta}-\bm{\theta}^{T} s\right]} \\ {\text { s.t. } W^{T} W=I_{c}, \bm{\theta}^{T} \bm{1}_{d}=1, \bm{\theta} \geq 0}\end{array}
\end{equation}

Eq.~\eqref{equ_solution_theta} is equivalent to the following function:
\begin{equation}
\label{equ_alm_theta}
\min\limits_{\bm{\theta}^{T} \bm{1}_{d}=1, \bm{\theta} \geq 0} \bm{\theta}^{T} Q \bm{\theta}-\bm{\theta}^{T} \bm{s}
\end{equation}
where\begin{equation}
\label{compute_Q_s}
\left\{
\begin{array}{l}
Q=\left( X{{H}^{T}}{{X}^{T}} \right)\circ \left( W{{W}^{T}} \right)+\lambda A\\
\bm{s} =diag\left( 2XHV{W}^{T} \right)
\end{array}
\right.
\end{equation}

To unravel the constrained optimization problem in Eq.~\eqref{equ_alm_theta}, we utilize the general augmented Lagrangian multiplier (ALM) method to further decompose Eq.~\eqref{equ_alm_theta} into the following subproblems:
\begin{equation}
\min\limits_{\bm{\theta}^{T} \bm{1}_{d}=1, \bm{v} \geq 0, \bm{v}=\bm{\theta}} \bm{\theta}^{T} Q \bm{\theta}-\bm{\theta}^{T} \bm{s}
\end{equation}

The augmented lagrangian of Eq.~\eqref{equ_alm_theta} is formulated as
\begin{equation}
\begin{aligned}
\label{equ_alm_app}
L\left(\bm{\theta}, \bm{v}, \mu, \bm{\delta}_{1}, \delta_{2}\right) = &\bm{\theta}^{T} Q \bm{\theta}-\bm{\theta}^{T} \bm{s}+\frac{\mu}{2}\left\|\bm{\theta}-\bm{v}+\frac{1}{\mu} \bm{\delta}_{1}\right\|_{F}^{2} \\
&+\frac{\mu}{2}\left(\bm{\theta}^{T} \mathbf{1}_{d}-1+\frac{1}{\mu} \delta_{2}\right)^{2} {\text { s.t. } \bm{v} \geq 0}
\end{aligned}
\end{equation}
where $\bm{v}$ and $\bm{\delta}_{1}$ are both column vectors, and $\mu$ is the Lagrangian multiplier. When $\bm{v}$ is fixed, Eq.~\eqref{equ_alm_app} can be rewritten as
\begin{equation}
\label{equ_alm_adj}
\min _{\bm{\theta}} \frac{1}{2} \bm{\theta}^{T} E \bm{\theta}-\bm{\theta}^{T} \bm{f}
\end{equation}
in which \begin{equation}\left\{\begin{array}{l}
E=2 Q+\mu I_{d}+\mu \bm{1}_{d} \bm{1}_{d}^{T} \\
\bm{f}=\mu \bm{v}+\mu \bm{1}_{d}-\delta_{2} \bm{1}_{d}-\bm{\delta}_{1}+\bm{s}
\end{array}\right.\end{equation}

We obtain the optimal solution of $\bm{\theta}$ is $\bm{\theta}=E^{-1} \bm{f}$.

When $\bm{\theta}$ is fixed, Eq.~\eqref{equ_alm_app} can be reformulated as the following:
\begin{equation}
\min _{\bm{v} \geq 0}\left\|\bm{v}-\left(\bm{\theta}+\frac{1}{\mu} \bm{\delta}_{1}\right)\right\|^{2}
\end{equation}

The optimal solution of $\bm{v}$ should be
\begin{equation}
\bm{v}=pos\left(\hat{\bm{\theta}}+\frac{1}{\mu} \bm{\delta}_{1}\right)
\end{equation}
where $pos\left(t\right)$ is a function that assigns 0 to each negative element of $t$.

\subsection{Update $V$ by fixing $\Theta$, $B$, and $W$}
With the fixed $\Theta$, $B$, and $W$, we set the derivatives w.r.t $V$ to zero. Considering $L$ is a symmetric matrix, we have

\begin{equation}
\label{partial_V}
2\left[H^{T}(V-X^{T} \Theta W) +\alpha (VB-Y)B^{T} + \eta LV\right] = 0
\end{equation}

Eq.~\eqref{partial_V} can be reformulated as:
\begin{equation}
\label{partial_V_trans}
(H^{T}+ \eta L)V + V(\alpha B B^{T}) = H^{T} X^{T} \Theta W + \alpha Y B^{T}
\end{equation}

Eq.~\eqref{partial_V_trans} is the matrix equation with the form of $ MV + VN = P $, where $ M = H^{T}+ \eta L $, $N = \alpha B B^{T}$, and $P = H^{T} X^{T} \Theta W + \alpha Y B^{T}$. $ MV + VN = P $ is the Sylvester equation. To solve the Sylvester equation, various practical methods have been successively proposed. Among them, the existed software library LAPACK and the lyap function in Matlab can be employed to derive the solution for $V$.

\subsection{Update $B$ by fixing $\Theta$, $V$, and $W$}

\begin{algorithm}[htbp]\footnotesize
\caption{Global Redundancy and Relevance Optimization in Orthogonal Regression (GRROOR)}
\label{GRROOR}
\begin{algorithmic}[1]
\Require The data matrix $X\in {\mathbb{R}^{\text{d}\times n}}$, the label matrix $Y\in {\mathbb{R}^{n\times k}}$. $p>1$, $\bm{\theta}_{i}=\frac{1}{d} \left(1\leq i \leq d \right)$, $\bm{v}=\bm{\theta}$, $\delta_2=0$, $u>0$, $\bm{\delta}_1 = \left( 0,0,\dots,0\right)^{T}\in {\mathbb{R}^{d\times 1}}$.
\Ensure Feature score vector $\bm{\theta}$.
\State Initial $\Theta \in {\mathbb{R}^{d\times d}}$ satisfying $\bm{\theta}^{T} \bm{1}_{d}=1$, and $\bm{\theta} \geq 0$. $H ={{I}_{\text{n}}}-\frac{1}{n}{{\bm{1}}_{n}}{{\bm{1}}_{n}}^{T}$. Initial $W$, $V$, and $B$ randomly.
\Repeat
\State Compute $ J=\Theta XH{{X}^{T}}{{\Theta }^{T}}$ and $M=\Theta XH{{V}^{T}}$
\State Update $W$ via GPI.
\Repeat
\State Update $Q$ and $\bm{s}$ via Eq.~\eqref{compute_Q_s};
\State Update $E$ by $E = 2Q+\mu I_d+\mu\bm{1}_{d}\bm{1}_{d}^{T}$;
\State Update $\bm{f}$ by $\bm{f} = \mu \bm{v} +\mu\bm{1}_{d}-\delta_{2}\bm{1}_{d}-\bm{\delta}_1+\bm{s}$;
\State Update $\bm{\theta}$ by $\bm{\theta}=E^{-1}\bm{f}$;
\State Update $\bm{v}$ by $\bm{v}=pos\left(\bm{\theta}+\frac{1}{\mu} \bm{\delta}_{1}\right)$;
\State Update $\bm{\delta}_1$ by $\bm{\delta}_1=\bm{\delta}_1+\mu\left(\bm{\theta}-\bm{v}\right)$;
\State Update $\delta_2$ by $\delta_2=\delta_2+\mu\left(\bm{\theta}^{T}\bm{1}_{d}-1\right)$;
\State Update $\mu$ by $\mu = p\mu$;
\Until{Convergence;}
\State Update $\Theta$  via $\Theta = diag(\bm{\theta})$;
\State Update $V$ by solving Eq.~\eqref{partial_V_trans};
\State Update $B$ by solving Eq.~\eqref{partial_B_trans};
\Until{Convergence;}
\State \Return $\bm{\theta}$ for multi-label feature selection.
\end{algorithmic}
\end{algorithm}

When $\Theta$, $V$, and $W$ are fixed, we can obtain the solution for $B$ by setting the derivatives w.r.t $B$ to zero, as follow:
\begin{equation}
\label{partial_B}
2\left[\alpha V^{T}(VB-Y)+\beta B R\right] = 0
\end{equation}

Eq.~\eqref{partial_B} can be converted to:
\begin{equation}
\label{partial_B_trans}
(\alpha V^{T} V)B + B (\beta R) = \alpha V^{T} Y
\end{equation}

The optimal solution to $B$ in Eq.~\eqref{partial_B_trans} can also be obtained by the existed software library LAPACK and the lyap function in Matlab.

Finally, the whole pseudocode for solving Eq.~\eqref{equ_framework} is shown in Algorithm~\ref{GRROOR}. The matrices $W$,$\Theta$, $V$, and $B$ are alternately updated until convergence. The feature score vector $\bm{\theta}$ is extracted from the final $\Theta$. The features are sorted on the basis of their corresponding values in $\bm{\theta}$. Lastly, the $m$ informative and non-redundant features with the top scores are selected.

\section{Experiment study} \label{Experimental Details}
In this section, the specific information regarding experimental data sets, comparing methods, performance metrics, and experiment setting will be illustrated. Then, extensive experiments are performed to validate the effectiveness of the proposed GRROOR method.

\subsection{Data set description}
The experimental studies are conducted on ten benchmark multi-label data sets\footnote{\url{http://www.uco.es/kdis/mllresources/}}, including Corel5k, Genbase, Image, Slashdot, Yeast, Entertainment, Education, Reference, Science, and Social data sets. Table~\ref{tab:datas} illustrates the details of each benchmark data set. We adopt the same train/test split approaches in Table~\ref{tab:datas} of these data sets to conduct experimental studies.

\begin{table}[htbp]\footnotesize
\setlength{\tabcolsep}{18pt}
\setlength{\abovecaptionskip}{0.cm}
\setlength{\belowcaptionskip}{-0.cm}
\caption{Information of multi-label data sets.}\label{tab:datas}
\begin{center}
{
\begin{tabular}{lccccc}
\hline\hline
Data set        & Training      & Test   & Instance   & Feature      & Label      \\\hline
Corel5k         & 4500          & 500    & 5000   & 499          & 374          \\
Genbase         & 463           & 199    & 662    & 1186         & 27         \\
Image           & 1000          & 1000   & 2000   & 294          & 5           \\
Slashdot        & 2546          & 1236   & 3782   & 1079         & 22             \\
Yeast           & 1499          & 918    & 2417   & 103          & 14          \\
Entertainment   & 2000          & 3000   & 5000   & 640          & 21              \\
Education       & 2000          & 3000   & 5000   & 550          & 33             \\
Reference       & 2000          & 3000   & 5000   & 793          & 33               \\
Science         & 2000          & 3000   & 5000   & 743          & 40              \\
Social          & 2000          & 3000   & 5000   & 1047         & 39              \\
\hline\hline
\end{tabular}}
\end{center}
\end{table}

\subsection{Comparing methods}
The proposed method is compared with nine state-of-the-art multi-label feature selection methods, including RFS \cite{nie2010RFS}, pairwise multi-label utility (PMU) \cite{lee2013feature}, feature selection based on information-theoretic feature ranking (FIMF) \cite{lee2015fast}, MIFS \cite{jian2016multi}, scalable criterion for a large label set (SCLS) \cite{lee2017scls}, MCLS \cite{huang2018manifold}, MFS-MCDM \cite{hashemi2020mfs}, global relevance and redundancy optimization (GRRO) \cite{zhang2020multi}, and SCMFS \cite{hu2020multi}. The parameters of each comparing algorithm are set as the corresponding reference suggested.

\subsection{Performance Metrics}
Six performance metrics are employed to compare the classification performance and redundant information removal performance from various aspects, including one feature redundancy evaluation metric redundancy, two label-based evaluation metrics macro-F1 and micro-F1, and three evaluation example-based metrics average precision, coverage, and hamming loss.

Let $\mathcal{U}=\left\{\left(\bm{x}_{.i}, \bm{y}_{i}\right) \mid 1 \leq i \leq n\right\}$ be a multi-label test set and $\mathrm{h}\left(\bm{x}_{.i}\right)$ be the learned multi-label set of the $i$-th instance $\bm{x}_{.i}$. The $\bm{x}_{.i}$. The definitions of the six metrics are described as follows.

(1) Hamming loss reflects the proportion of mislabeled labels. $\oplus$ is a symmetric difference operator.

\begin{equation}
H L =\frac{1}{n} \sum_{i=1}^{n} \frac{1}{k}\left|\mathrm{h}\left(\bm{x}_{.i}\right) \oplus \bm{y}_{i}\right|_{1}
\end{equation}

(2) Coverage computes the number of steps required to find all the ground-truth labels of one instance from the label ranking sequence.

\begin{equation}
C V =\frac{1}{k}\left(\frac{1}{n} \sum_{i=1}^{n} \max _{l_{r} \in \bm{y}_{i}} \operatorname{rank}\left(\bm{x}_{.i}, l_{r}\right)-1\right)
\end{equation}

(3) Average precision is used to calculate the average proportion of related labels higher than a given label in the label ranking list.
\begin{equation}
A P =\frac{1}{n} \sum_{i=1}^{n} \frac{1}{\left|\bm{y}_{i}\right|} \sum_{l \in \bm{y}_{i}} \frac{\left|L_{i}=\left\{l_{j} \mid \operatorname{rank}\left(\bm{x}_{.i}, l_{j}\right) \leq \operatorname{rank}\left(\bm{x}_{.i}, l_{r}\right)\right\}\right|}{\operatorname{rank}\left(\bm{x}_{.i}, l_{r}\right)}
\end{equation}

(4) Macro-F1 measures the average F-measure value over all labels to evaluate the label set prediction performance of a classifier.
\begin{equation}
\operatorname{Macro-F1}=\frac{1}{k} \sum_{j=1}^{k} \frac{2 \sum_{i=1}^{n} y_{i j} \mathrm{h}_{j}\left(\bm{x}_{.i}\right)}{\sum_{i=1}^{n} y_{i j}+\sum_{i=1}^{n} \mathrm{h}_{j}\left(\bm{x}_{.i}\right)}
\end{equation}

(5) Micro-F1 is an average of F-measure values on the prediction matrix to measure the label set prediction performance of a classifier.
\begin{equation}
\operatorname{Micro-F1} =\frac{2 \sum_{i=1}^{n}\left|\mathrm{h}\left(\bm{x}_{.i}\right) \cap \bm{y}_{i}\right|_{1}}{\sum_{i=1}^{n}\left|\bm{y}_{i}\right|_{1}+\sum_{i=1}^{n}\left|\mathrm{h}\left(\bm{x}_{.i}\right)\right|_{1}}
\end{equation}

(6) Redundancy is used to evaluate the redundant information among the selected feature subset. $m$ is the number of selected features in the feature subset $G$ and $A_{i,j}$ is the squared cosine similarity of the features $\bm{x}_{i}$ and $\bm{x}_{j}$.
\begin{equation}
\operatorname{Redundancy}(G)=\frac{1}{m(m-1)} \sum_{\bm{f}_{i}, \bm{f}_{j} \in G, i \neq j} A_{i, j}
\end{equation}

In terms of coverage, redundancy, and hamming loss, the value is expected as small as possible. While in terms of macro-F1, micro-F1, and average precision, a larger value brings to better multi-label classification results.

\begin{sidewaystable*}[htbp]\scriptsize
\setlength{\tabcolsep}{10pt}
\setlength{\abovecaptionskip}{0.cm}
\setlength{\belowcaptionskip}{-0.cm}
\caption{Comparison results of multi-label feature selection methods in terms of redundancy, coverage, and hamming loss}\label{tab:per1}
 \begin{center}
 {
\begin{tabular}{lcccccccccccc}
\hline\hline
\multicolumn{1}{l}{\multirow{2}*{Data sets}}         & \multicolumn{10}{c}{\multirow{1}*{\textbf{Redundancy} $\downarrow$}} \\ \cline{2-11}
                                                     & Yeast	& Social	& Slashdot	& Science	& Reference	 & Image	  & Genbase	  & Entertainment  & Corel\_5k	  & Education \\ \hline
RFS                                 & 0.1353 & 0.1864 & 0.0883 & 0.1639 & 0.2006 & 0.8640 & 0.0009 & 0.6694 & 0.0652 & 0.2488\\
MCLS                         & 0.1570 & 0.1886 & 0.0740 & 0.1803 & 0.1592 & 0.7189 & 0.0025 & 0.1787 & 0.0017 & 0.2212\\
PMU                            & 0.1382 & 0.1821 & 0.1159 & 0.1690 & 0.1392 & 0.4912 & 0.0213 & 0.1491 & 0.0647 & 0.1784\\
SCLS                               & 0.1382 & 0.1821 & 0.1053 & 0.1531 & 0.1509 & 0.4913 & 0.0012 & 0.1631 & 0.0647 & 0.1926\\
MIFS                             & 0.1273 & 0.1965 & 0.0694 & 0.1636 & 0.1720  & 0.4736 & 0.1383 & 0.1755 & 0.0648 & 0.1862\\
SCMFS                              & 0.1359 & 0.1821 & 0.1282 & 0.1774 & 0.2587 & 0.5407 & 0.0007 & 0.2739 & 0.0647 & 0.2015\\
FIMF                               & 0.1286 & 0.1958 & 0.0621 & 0.1714 & 0.1820  & 0.4334 & 0.0824 & 0.1727 & \textbf{0.0009} & 0.1655\\
GRRO                            & 0.1620 & 0.1856 & 0.0896 & 0.1530 & 0.1914 & 0.4094 & 0.1220 & 0.3136 & 0.0142 & 0.1779\\
MFS\_MCDM                       & 0.1668 & 0.1488 & 0.0794 & 0.1559 & 0.1617 & 0.4031 & 0.0010 & 0.1388 & 0.0511 & \textbf{0.1394}\\
\textbf{GRROOR(our)}                                & \textbf{0.1208} 	& \textbf{0.1312} 	&\textbf{0.0518} 	& \textbf{0.1189} 	& \textbf{0.1312} 	& \textbf{0.2831} 	& \textbf{0.0005} 	& \textbf{0.1205} 	& 0.0011  & 0.1402\\
\hline\hline
\multicolumn{1}{l}{\multirow{2}*{Data sets}}         & \multicolumn{10}{c}{\multirow{1}*{\textbf{Coverage} $\downarrow$}} \\  \cline{2-11}
                                                     & Yeast	& Social	& Slashdot	& Science	& Reference	 & Image	  & Genbase	  & Entertainment  & Corel\_5k	  & Education \\ \hline
RFS                                 &6.6696 &3.8141 &5.2259 &7.5396 &3.8519 &1.8489 &0.7495 &3.6903 &118.6296 &5.5078\\
MCLS                         &6.6833 &4.1310 &5.0966 &7.5561 &3.5955 &1.5432 &1.4528 &3.7621 &119.3255 &5.4262\\
PMU                             &6.5941 &3.7004 &4.4943 &7.3058 &3.6472 &1.2151 &0.8161 &3.6237 &120.0134 &5.7285\\
SCLS                               &6.5170 &3.6820 &4.3083 &6.9508 &3.4597 &1.1895 &0.7547 &3.3191 &118.6864 &5.4702\\
MIFS                             &6.5633 &3.7238 &3.8167 &7.0752 &3.4699 &1.5149 &0.6162 &3.4183 &117.2359 &4.9113\\
SCMFS                              &6.5201 &3.4735 &3.7593 &6.7337 &3.2237 &1.0413 &0.6018 &3.2001 & 108.9085 &4.7531\\
FIMF                               &6.6960 &3.6903 &4.2673 &7.1885 &3.5620 &1.3325 &0.8412 &3.6530 &120.0582 &5.5977\\
GRRO                            &6.5744 &3.6923 &4.2551 &6.8652 &3.5126 &1.2286 &0.7994 &3.3758 &118.1895 &5.4977\\
MFS\_MCDM                       &6.7852 &3.4983 &3.8948 &7.1449 &3.6655 &1.2511 &0.8178 &3.4632 &118.6772 &5.1183\\
\textbf{GRROOR(our)}                                &\textbf{6.4981} & \textbf{3.4060} & \textbf{3.6538} & \textbf{6.6094} & \textbf{3.1994} & \textbf{1.0280} & \textbf{0.5927} & \textbf{3.1368} &\textbf{108.7914} & \textbf{3.8607}\\
\hline\hline
\multicolumn{1}{l}{\multirow{2}*{Data sets}}         & \multicolumn{10}{c}{\multirow{1}*{\textbf{Hamming loss} $\downarrow$}} \\  \cline{2-11}
                                                     & Yeast	& Social	& Slashdot	& Science	& Reference	 & Image	  & Genbase	  & Entertainment  & Corel\_5k	  & Education \\ \hline
RFS                                 & 0.2137 & 0.0313 & 0.0539 & 0.0356 & 0.0356 & 0.2411 & 0.0080 & 0.0662 & 0.0097 & 0.0441\\
MCLS                         & 0.2209 & 0.0331 & 0.0537 & 0.0356 & 0.0337 & 0.2292 & 0.0268 & 0.0672 & 0.0094 & 0.0442\\
PMU                             & 0.2117 & 0.0254 & 0.0510 & 0.0355 & 0.0302 & 0.2065 & 0.0094 & 0.0650 & 0.0094 & 0.0415\\
SCLS                               & 0.2088 & 0.0245 & 0.0492 & 0.0340 & 0.0288 & 0.2034 & 0.0070 & 0.0591 & 0.0095 & 0.0412\\
MIFS                             & 0.2055 & 0.0246 & 0.0456 & 0.0347 & 0.0302 & 0.2227 & 0.0054 & 0.0622 & 0.0094 & 0.0417\\
SCMFS                              & 0.2051 & 0.0236 & 0.0455 & 0.0342 & 0.0284 & 0.1793 & 0.0048 & 0.0593 & 0.0094 & 0.0411\\
FIMF                               & 0.2140 & 0.0246 & 0.0499 & 0.0345 & 0.0296 & 0.2090 & 0.0070 & 0.0590 & 0.0095 & 0.0409\\
GRRO                            & 0.2082 & 0.0240 & 0.0490 & 0.0337 & 0.0285 & 0.2090 & 0.0070 & 0.0590 & 0.0095 & 0.0409\\
MFS\_MCDM                       & 0.2263 & 0.0240 & 0.0489 & 0.0341 & 0.0292 & 0.2063 & 0.0069 & 0.0611 & 0.0094 & 0.0428\\
\textbf{GRROOR(our)}                                & \textbf{0.2046} 	& \textbf{0.0230} 	& \textbf{0.0450} 	& \textbf{0.0337} 	& \textbf{0.0281} 	& \textbf{0.1775} 	& \textbf{0.0043} 	& \textbf{0.0583} 	& \textbf{0.0093} 	& \textbf{0.0407} \\
\hline\hline
\end{tabular}}
\end{center}
\end{sidewaystable*}

\begin{sidewaystable*}[htbp]
\scriptsize
\setlength{\tabcolsep}{10pt}
\setlength{\abovecaptionskip}{0.cm}
\setlength{\belowcaptionskip}{-0.cm}
\caption{Comparison results of multi-label feature selection methods in terms of average precision, macro-F1, and micro-F1}\label{tab:per2}
 \begin{center}
 {
\begin{tabular}{lcccccccccccc}
\hline\hline
\multicolumn{1}{l}{\multirow{2}*{Data sets}}         & \multicolumn{10}{c}{\multirow{1}*{\textbf{Average precision} $\uparrow$}} \\  \cline{2-11}
                                                     & Yeast	& Social	& Slashdot	& Science	& Reference	 & Image	  & Genbase	  & Entertainment  & Corel\_5k	  & Education \\ \hline
RFS                                 &0.7364    &0.6349     &0.3624     &0.4017     &0.5751      &0.5721      &0.9543     &0.5037          &0.2195        &0.4047 \\
MCLS                         &0.7253    &0.5971     &0.3855     &0.3961     &0.5978      &0.6473      &0.8206     &0.4942          &0.2272        &0.4089 \\
PMU                             &0.7417    &0.6974     &0.4406     &0.4303     &0.5988      &0.7265      &0.9551     &0.5138          &0.2153        &0.4861\\
SCLS                               &0.7460    &0.7073     &0.4729     &0.4677     &0.6200      &0.7349      &0.9584     &0.5754          &0.2286        &0.5120\\
MIFS                             &0.7454    &0.6918     &0.5543     &0.4578     &0.6030      &0.6432      &0.9855     &0.5570          &0.2332        &0.4880\\
SCMFS                              &0.7451    &0.7209     &0.5612     &0.4806     &0.6419      &0.7720      &0.9858     &0.5931          &\textbf{0.2847} &0.5007\\
FIMF                               &0.7360    &0.7039     &0.4659     &0.4542     &0.6064      &0.6882      &0.9505     &0.5111          &0.2157         &0.4923\\
GRRO                            &0.7432    &0.7124     &0.4767     &0.4747     &0.6198      &0.7240      &0.9517     &0.5713          &0.2336         &0.5106\\
MFS\_MCDM                       &0.7181    &0.7173     &0.5187     &0.4643     &0.6063      &0.7187      &0.9522     &0.5432          &0.2293         &0.4566\\
\textbf{GRROOR(our)}                                &\textbf{0.7481} &\textbf{0.7270} &\textbf{0.5733} &\textbf{0.4942} &\textbf{0.6464} &\textbf{0.7780} &\textbf{0.9876} &\textbf{0.6066} &0.2839 &\textbf{0.5588}\\
\hline\hline
\multicolumn{1}{l}{\multirow{2}*{Data sets}}         & \multicolumn{10}{c}{\multirow{1}*{\textbf{Macro-F1} $\uparrow$}} \\ \cline{2-11}
                                                     & Yeast	& Social	& Slashdot	& Science	& Reference	 & Image	  & Genbase	  & Entertainment  & Corel\_5k	  & Education \\ \hline
RFS                                 & 0.2856 & 0.0802 & 0.0979 & 0.0065 & 0.0756 & 0.0771 & 0.6126 & 0.0330 & 0.2979 & 0.0653\\
MCLS                         & 0.2422 & 0.0535 & 0.1055 & 0.0050 & 0.0964  & 0.2195 & 0.3554 & 0.0188 & 0.2994 & 0.0643\\
PMU                             & 0.3049 & 0.1138 & 0.1718 & 0.0163 & 0.0399  & 0.4281 & 0.5393 & 0.0641 & 0.2968 & 0.1274\\
SCLS                               & \textbf{0.3320} & 0.1213 & 0.2126 & 0.0486 & 0.1200  & 0.4486 & 0.5663 & 0.1183 & 0.2981 & 0.1272\\
MIFS                             & 0.3172 & 0.0928 & 0.2284 & 0.0292 & 0.0994 & 0.2047 & 0.5992 & 0.0806 & 0.2986 & 0.1214\\
SCMFS                              & 0.3242 & 0.1248 & 0.2357 & 0.0474 & 0.1355 & 0.5244 & 0.6301 & 0.1311 & 0.3058 & 0.1282\\
FIMF                               & 0.3008 & 0.1099 & 0.2024 & 0.0387 & 0.1040 & 0.3170 & 0.5903 & 0.0642 & 0.2968 & 0.1304\\
GRRO                            & 0.3205 & 0.1190 & 0.2193 & 0.0534 & 0.1236 & 0.4270 & 0.5858 & 0.1102 & 0.2979 & 0.1306\\
MFS\_MCDM                       & 0.2085 & 0.1231 & 0.2393 & 0.0481 & 0.1111 & 0.3912 & 0.5977 & 0.1115 & 0.3007 & 0.0982\\
\textbf{GRROOR(our)}                                & 0.3276 & \textbf{0.1340} 	& \textbf{0.2484} 	& \textbf{0.0627} 	& \textbf{0.1471} 	 & \textbf{0.5402} 	& \textbf{0.6840} 	& \textbf{0.1420} 	& \textbf{0.3065}  & \textbf{0.1340}\\
\hline\hline
\multicolumn{1}{l}{\multirow{2}*{Data sets}}         & \multicolumn{10}{c}{\multirow{1}*{\textbf{Micro-F1} $\uparrow$}} \\  \cline{2-11}
                                                     & Yeast	  & Social	& Slashdot	& Science	& Reference	 & Image	  & Genbase	  & Entertainment  & Corel\_5k	  & Education \\ \hline
RFS                                 &0.5842 &0.1993 &0.0208 &0.0115 &0.1272 &0.0877 &0.8970 &0.0627 &0.0137 &0.0285\\
MCLS                         &0.5476 &0.0156 &0.0488 &0.0074 &0.2412  &0.2492 &0.6362 &0.0317 &0.0080 &0.0368\\
PMU                             &0.5949 &0.4379 &0.1338 &0.0407 &0.3265  &0.4355 &0.8778 &0.1085 &0.0003 &0.1227\\
SCLS                               &0.6085 &0.4813 &0.2220 &0.1502 &0.3693 &0.4564 &0.9053 &0.2806 &0.0050 &0.1891\\
MIFS                             &0.6119 &0.4632 &0.3471 &0.0970 &0.3173 &0.2695 &0.9380 &0.2081 &0.0075 &0.1479\\
SCMFS                              &0.6125 &0.5045 &0.3591 &0.1330 &0.4009 &0.5277 &0.9450 &0.3004 &0.0356 &0.1793\\
FIMF                               &0.5976 &0.4740 &0.1990 &0.1148 &0.3289  &0.3320 &0.8948 &0.1217 &0.0010 &0.1276\\
GRRO                            &0.6059 &0.5040 &0.2276 &0.1659 &0.3796 &0.4394 &0.9065 &0.2866 &0.0064 &0.1967\\
MFS\_MCDM                      &0.5288 &0.5026 &0.3159 &0.1461 &0.3642 &0.4047 &0.9056 &0.2430 &0.0290 &0.1207\\
\textbf{GRROOR(our)}                                &\textbf{0.6166} &\textbf{0.5322} &\textbf{0.3721} &\textbf{0.1829} &\textbf{0.4267} &\textbf{0.5402} &\textbf{0.9500} &\textbf{0.3214} &\textbf{0.0419} &\textbf{0.2254}\\
\hline\hline
\end{tabular}}
\end{center}
\end{sidewaystable*}

\subsection{Experiment setting}
Multi-label k-Nearest Neighbor (ML-KNN) \cite{ZHANG2007MLKNN} is employed to measure the performance of feature selection methods. The neighbor number and smooth are set to 10 and 1, respectively. We record the classification performance by changing the size of the selected feature subset from 1 to 50 with step 1. The experiments are repeated 10 times to avoid bias. The average and standard deviation results with 50 groups of feature subsets are used to compare.

For the proposed method, we tune the tradeoff parameters ($\lambda$, $\eta$, and $\beta$) with grid-search strategy in the range of $\left\{10^{-3},10^{-2},10^{-1},0.2,0.4,0.6,0.8,10,100\right\}$, and $c$ in $\left\{2,0.25k,0.5k,0.75k,k\right\}$. To avoid the influence of the value of the tradeoff parameter $\alpha$ on the two items $tr\left(V^{T} L V\right)$ and $tr\left(\mathrm{RB^{T}B}\right)$, the value of $\alpha$ is set to 1. The value of $\sigma^{2}$ and $p$ in the definition of affinity graph $S$ is set to 1 and 5 to model the local geometry structure in the data space $X$. The parameters of each comparing algorithm are set as the corresponding reference suggested. We adopt the average classification result (ACR) as an indicator for seeking the optimal parameters \cite{zhang2020multi}. For ACR, the smaller the value, the better the performance. The definition of ACR is:

\begin{equation}
ACR(\mathrm { para })=\sum_{i=1}^{30}\left(H L_{i}((\mathrm{h}, \mathcal{U}))+R L_{i}(\mathrm{h}, \mathcal{U})\right)
\end{equation}
where $\mathrm { para }$ represents the collection of parameters and $i$ denotes the number of selected top-$i$ features.

\subsection{Experimental Results and Discussion}\label{Experimental Results and Discussion}

\begin{figure*}[!htb]
\centering
\subfigure[]{\label{Slashdot_Redu}\includegraphics[width=0.3\textwidth]{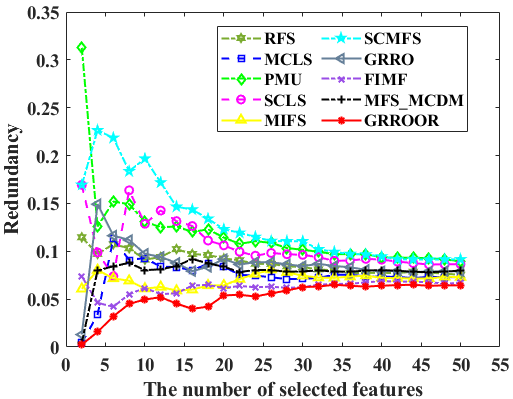}}
\hspace{0.0cm}
\subfigure[]{\label{Slashdot_CV}\includegraphics[width=0.3\textwidth]{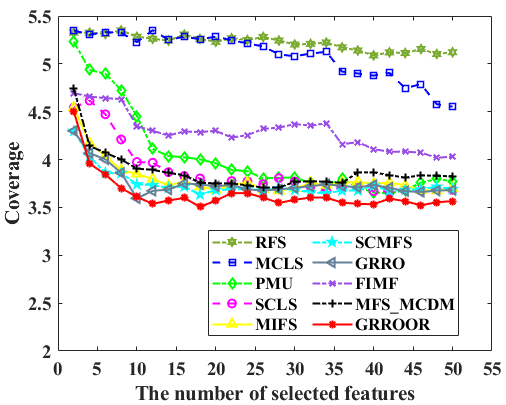}}
\hspace{0.0cm}
\subfigure[]{\label{Slashdot_HL}\includegraphics[width=0.3\textwidth]{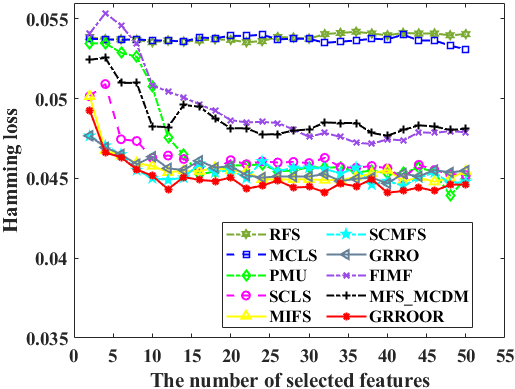}}
\hspace{0.0cm}
\subfigure[]{\label{Slashdot_AP}\includegraphics[width=0.3\textwidth]{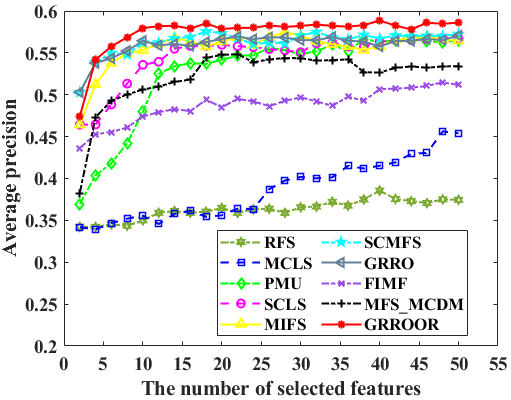}}
\hspace{0.0cm}
\subfigure[]{\label{Slashdot_MA}\includegraphics[width=0.3\textwidth]{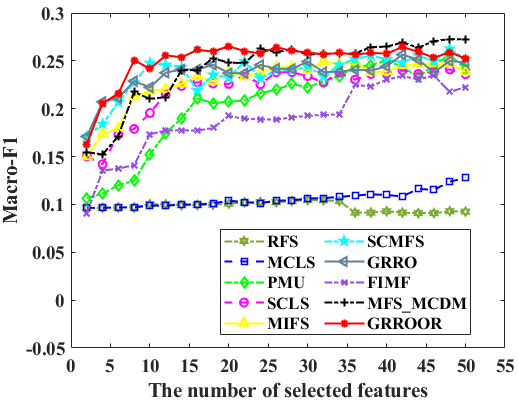}}
\hspace{0.0cm}
\subfigure[]{\label{Slashdot_MI}\includegraphics[width=0.3\textwidth]{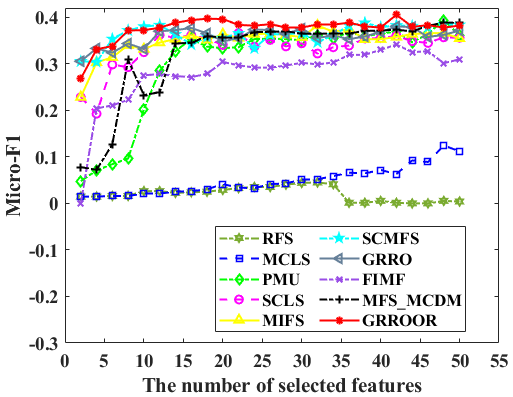}}
\caption{Multi-label classification performance with different number of selected features on the Slashdot data set: (a) Redundancy; (b) Coverage; (c) Hamming loss; (d) Average precision; (e) Macro-F1; (f)Micro-F1.}\label{Results_index}
\end{figure*}

In this section, the proposed GRROOR method is compared with nine other comparison algorithms in terms of six performance metrics.Table~\ref{tab:per1} and Table~\ref{tab:per2} report the average for the different number of selected features, and the best results in all the evaluation metrics are shown in bold. It should be noted that, for each evaluation measure, $\downarrow$ illustrates the smaller the better and $\uparrow$ implies the larger the better. As shown in Table~\ref{tab:per1} and Table~\ref{tab:per2}, we can observe: 1) the GRROOR method can achieve optimal average classification performances at least on eight data sets; 2) the GRROOR method can achieve sub-optimal classification performances among all the comparison methods on the Corel\_5k data set for two evaluation metrics (redundancy and average precision) and on the Yeast data set for macro\_F1.

\begin{table}[htbp]\footnotesize
\setlength{\tabcolsep}{18pt}
\setlength{\abovecaptionskip}{0.cm}
\setlength{\belowcaptionskip}{-0.cm}
\caption{Friedman test results (10 comparing algorithms, 10 data sets, significance level $\alpha = 0.05$)}\label{tab:ccc}
\begin{center}
{
\begin{tabular}{lcc}
\hline\hline
Evaluation metric     & $F_{F}$             & Critical value  \\ \hline
Redundancy            & 4.0825              & \multicolumn{1}{c}{\multirow{6}*{$\approx$ 1.998}}             \\
Coverage              & 22.5421             &            \\
Hamming loss          & 24.3333             &            \\
Average precision     & 28.3491             &           \\
Macro-F1              & 20.1005             &            \\
Micro-F1              & 26.1230             &            \\
\hline\hline
\end{tabular}}
\end{center}
\end{table}

\begin{figure*}[!t]
\centering
\subfigure[]{\label{Redu_Neme}\includegraphics[width=0.3\textwidth]{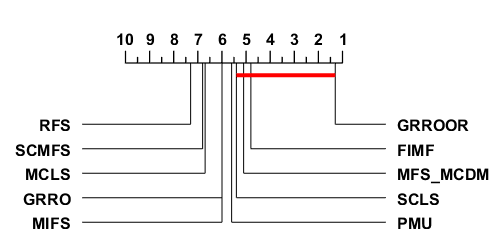}}
\hspace{0.0cm}
\subfigure[]{\label{CV_Neme}\includegraphics[width=0.3\textwidth]{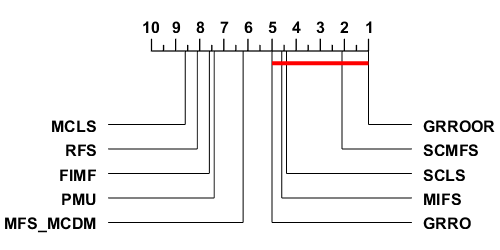}}
\hspace{0.0cm}
\subfigure[]{\label{HL_Neme}\includegraphics[width=0.3\textwidth]{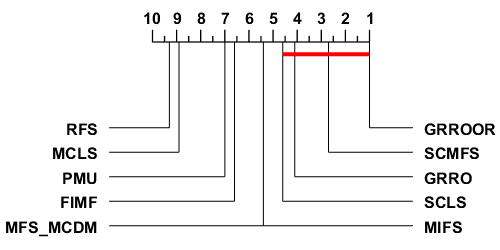}}
\hspace{0.0cm}
\subfigure[]{\label{AP_Neme}\includegraphics[width=0.3\textwidth]{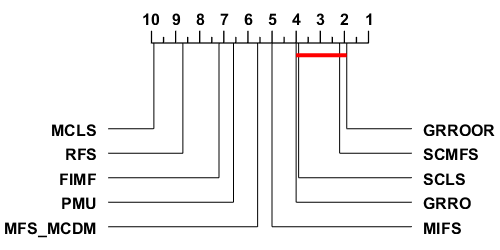}}
\hspace{0.0cm}
\subfigure[]{\label{MA_Neme}\includegraphics[width=0.3\textwidth]{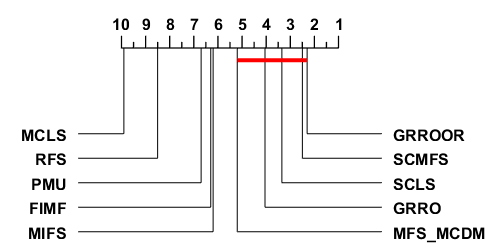}}
\hspace{0.0cm}
\subfigure[]{\label{MI_Neme}\includegraphics[width=0.3\textwidth]{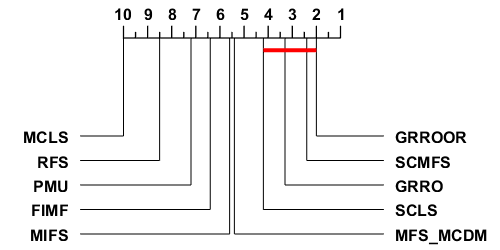}}
\caption{The Nemenyi test results ($C D = 4.2841$, $\alpha =0.05$): (a) Redundancy; (b) Coverage; (c) Hamming loss; (d) Average precision; (e) Macro-F1; (f) Micro-F1.} \label{Results_neme}
\end{figure*}

\begin{figure*}[!t]
\centering
\subfigure[]{\label{bar3_beta_lambda}\includegraphics[width=0.35\textwidth]{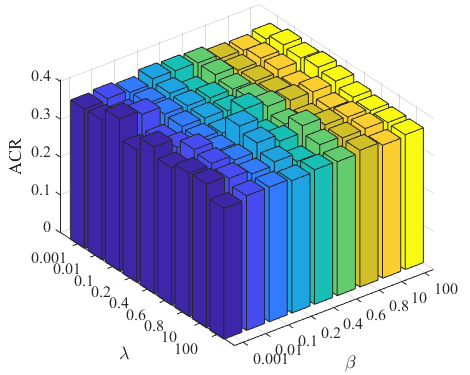}}
\hspace{0.0cm}
\subfigure[]{\label{bar3_beta_eta}\includegraphics[width=0.35\textwidth]{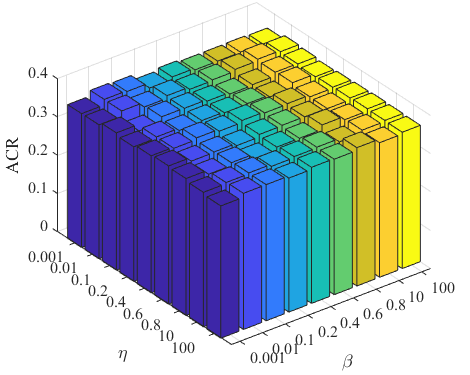}}
\hspace{0.0cm}
\subfigure[]{\label{bar3_lambda_eta}\includegraphics[width=0.35\textwidth]{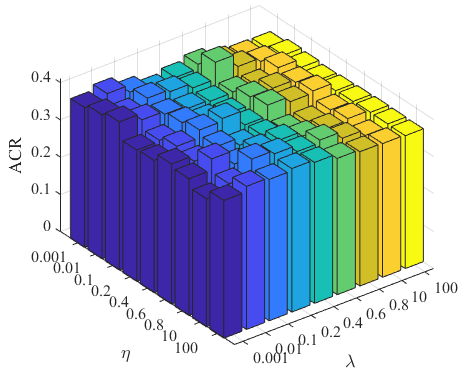}}
\hspace{0.0cm}
\subfigure[]{\label{Conv_slashdot}\includegraphics[width=0.35\textwidth]{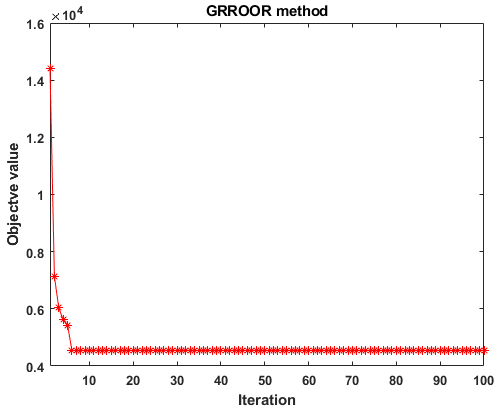}}
\caption{The parameter sensitivity (under the varying $\lambda$, $\beta$, and $\eta$) and convergence of GRROOR on the Image data set.}\label{Results_bar3}
\end{figure*}

To graphically show the performance of ten multi-label feature selection methods, the results of the Slashdot data set are chosen. Fig.~\ref{Results_index} shows the classification performance of different feature selection methods on the Slashdot data set. For each subfigure, the horizontal axis represents the number of features selected by the multi-label feature selection methods and the vertical axis represents the values of the performance metrics. We can observe that the values in terms of three evaluation metrics (macro-F1, micro-F1, and average precision) tend to increase and then begin to degrade slightly or keep stable as the number of selected features increased. The above results indicate that the feature selection step is necessary to select discriminative features and remove redundant or noisy features from the original features of the ten data sets. Additionally, the GRROOR method could obtain stable performance of all the performance metrics significantly faster and maintain it more stably, which demonstrates that the GRROOR method achieves better classification performances than other compared methods.

To further analyze the relative performance between GRROOR and comparing methods. The Friedman test is employed as the favorable statistical significance test for the classification performance comparison of ten methods. Table~\ref{tab:ccc} shows the Friedman statistics of each evaluation measure and the critical value at significance level $\alpha = 0.05$, which indicates that the null hypothesis is rejected and the multi-label feature selection performance of ten methods has a significant difference. To complete the performance comparison, the Nemenyi test is then introduced for certain post-hoc test,  where the GRROOR method is regarded as the control method. For the Nemenyi test, the critical difference ($C D$), employed to control the family-wise error rate, can be calculated as follows:
\begin{equation}
C D=q_{\alpha} \sqrt{\frac{nc(nc+1)}{6 nd}}
\end{equation}
where $nc$ and $nd$ denote the number of methods and datasets. The $q_{\alpha}$ is 3.164 at $\alpha = 0.05$. $C D$ can be computed as $CD = 4.2841$ ($nc$ = 10, $nd$ = 10).

Fig.~\ref{Results_neme} shows the Nemenyi test results under six evaluation measures. If the average rank value of the control method is within one $C D$ to those of the compared methods, the methods are connected by a red line, which shows the difference between the comparing method and the GRROOR method is not so obvious. Otherwise, the comparing method is unconnected with the control method. As can be seen from Fig.~\ref{Results_neme}, although the performance of GRROOR is not significantly different from those of the comparing method on all the evaluation measures, GRROOR ranks 1st among all the methods on each performance metric. Hence, the results in Fig.~\ref{Results_neme} illustrate that GRROOR can obtain highly competitive performance against all the compared methods.

\subsection{Computational complexity analysis}
Here, $\mathcal{O}$ represents the computational cost of the algorithm. To reduce the computational cost of calculating $W$ in the GRROOR algorithm, direct calculation of $JW$ is employed instead of calculating the matrix $J$ and then multiplying by the matrix $W$. The GRROOR algorithm costs $\mathcal{O}\left(dkn\right)$ to compute the $JW$. The computational complexity of calculating $\Theta$ is also $\mathcal{O}\left(dkn\right)$. The GRROOR algorithm requires $\mathcal{O}\left(n^{3}\right)$ to compute $V$ and $\mathcal{O}\left(c^{3}\right)$ to calculate $B$. Finally, the total computational complexity of the proposed method is $\mathcal{O}\left(dkn+n^{3}+c^{3}\right)$.

\subsection{Parameter sensitivity analysis and convergence demonstration}
In GRROOR, three parameters $\lambda$, $\beta$, and $\eta$ should be set in advance. To study the parameter sensitivity of our proposed algorithm, we conduct an experiment to evaluate the influence of the three parameters and report the performance variances. We tune two parameters while fixing the other parameter as 100. Due to space limitations, we only show the ACR results of the Image data set with the top 50 ranked features in Fig.4(a-c). As shown in Fig.4(a-c), the ACR changes when different pairs of parameters are employed, and the optimal performance is obtained with moderate $\lambda$ and $\beta$. Therefore, the performance of GRROOR is sensitive to the values of control parameters.

To study the convergence of our iterative optimization algorithm, the convergence learning curve on the Image data set is shown in Fig.~\ref{Conv_slashdot}. The parameters $\lambda$, $\beta$, and $\eta$ are set to 10. As shown in Fig.~\ref{Conv_slashdot}, the objective function values of GRROOR monotonically decline at the few iterations and converge within 6 iterations, which demonstrates the effectiveness and stability of GRROOR.

\section{Conclusions} \label{Conclusions}
The state-of-the-art LSR-based multi-label feature selection methods usually cannot preserve sufficient discriminative information in the multi-label data. To resolve the problem, in this paper, we propose an embedded multi-label feature selection framework to select discriminative and non-redundant features via concurrently merging global redundancy and relevance optimization in the orthogonal regression with feature weighting. Compared with LSR based methods, the GRROOR adopts orthogonal regression to retain more local structural information of multi-label data, which is beneficial to capturing the relationship between the features and labels. Additionally, the GRROOR framework could simultaneously exploit feature redundancy and label relevance from a global view.

An efficient iterative optimization algorithm is proposed to solve the unbalanced orthogonal Procrustes problem in the objective function of the GRROOR method. Finally, GRROOR is compared with nine multi-label feature selection methods on ten multi-label data sets in terms of six performance metrics. The experimental results validate the superior performance of GRROOR.

Nevertheless, in contrast with filter methods, GRROOR often requires higher computational time cost. The computational complexity of the GRROOR method consists of Cubic order of $n$. It is worth mentioning that the heavy computational cost may limit the application of GRROOR in the real scene with extremely large sample sizes ($n$).

Multi-label feature selection with missing labels has attracted extensive attention in the field of pattern recognition. Due to the limitation of LSR mentioned above, the existing LSR-based multi-label feature selection could not accurately model the complex relationship between the features and incomplete labels. In future work, we will further investigate orthogonal regression based multi-label feature selection framework under the circumstance of missing labels. Local and global label relevancy could be used simultaneously to recover the missing labels.

\section{Acknowledge} \label{Acknowledge}
This work is funded by: the Special Fund for Research on National Major Research Instruments of the Nature Science Foundation of China under Grant No.62227801, and the Beijing Natural Science Foundation under Grant No.4212037, the China Postdoctoral Science Foundation under Grant
No.2022M720332, and the Beijing Postdoctoral Science Foundation under Grant No.2023-zz-85.


%




\section{Reference} \label{Reference}

\bibliographystyle{elsarticle-num}
\bibliography{IEEErefpr}

\end{document}